\documentclass[letterpaper]{article} 
\usepackage{aaai2026}  
\usepackage{times}  
\usepackage{helvet}  
\usepackage{courier}  
\usepackage[hyphens]{url}  
\usepackage{graphicx} 
\usepackage{natbib}  
\usepackage{caption} 
\usepackage{algorithm}
\usepackage{booktabs}
\usepackage{multirow}
\usepackage{amsmath}
\usepackage[noend]{algpseudocode}
\usepackage{subcaption}
\usepackage{xcolor}

\usepackage{newfloat}
\usepackage{listings}
\DeclareCaptionStyle{ruled}{labelfont=normalfont,labelsep=colon,strut=off} 
\floatstyle{ruled}
\newfloat{listing}{tb}{lst}{}
\floatname{listing}{Listing}
\title{A General Highly Accurate Online Planning Method Integrating Large Language Models into Nested Rollout Policy Adaptation for Dialogue Tasks}
\author {
     Hui Wang\textsuperscript{\rm 1, \rm 2}, 
     Fafa Zhang\textsuperscript{\rm 1, \rm 2}, 
     Xiaoyu Zhang\textsuperscript{\rm 1, \rm 2}, 
     Chaoxu Mu\textsuperscript{\rm 1, \rm 2, \rm 3}\thanks{Corresponding author: Chaoxu Mu.}
}
\affiliations {
    \textsuperscript{\rm 1}School of Artificial Intelligence, Anhui University\\
    \textsuperscript{\rm 2}Anhui Provincial Key Laboratory of Security Artificial Intelligence, Anhui University\\
    \textsuperscript{\rm 3}Pengcheng Laboratory, Shenzhen, China\\
    \{h.wang.13, zhangxiaoyu\}@ahu.edu.cn, wa23301160@stu.ahu.edu.cn, cxmu@tju.edu.cn
}

\begin{document}

\maketitle

\begin{abstract}
 In goal-oriented dialogue tasks, the main challenge is to steer the interaction towards a given goal within a limited number of turns. Existing approaches either rely on elaborate prompt engineering, whose effectiveness is heavily dependent on human experience, or integrate policy networks and pre-trained policy models, which are usually difficult to adapt to new dialogue scenarios and costly to train. Therefore, in this paper, we present Nested Rollout Policy Adaptation for Goal-oriented Dialogue (NRPA-GD), a novel dialogue policy planning method that completely avoids specific model training by utilizing a Large Language Model~(LLM) to simulate behaviors of user and system at the same time. Specifically, NRPA-GD constructs a complete evaluation mechanism for dialogue trajectories and employs an optimization framework of nested Monte Carlo simulation and policy self-adaptation to dynamically adjust policies during the dialogue process. The experimental results on four typical goal-oriented dialogue datasets show that NRPA-GD outperforms both existing prompt engineering and specifically pre-trained model-based methods. Impressively, NRPA-GD surpasses ChatGPT and pre-trained policy models with only a 0.6-billion-parameter LLM. The proposed approach further demonstrates the advantages and novelty of employing planning methods on LLMs to solve practical planning tasks.
\end{abstract}


\section{Introduction}

With the emergence of Large Language Models~(LLMs)~\cite{LLM1} represented by ChatGPT~\cite{LLM2}, in the field of Natural Language Processing~(NLP)~\cite{NLP}, there are many key technological breakthroughs for dialogue tasks~\cite{dialog}. Currently, LLM integration methods have significantly improved overall performance for goal-oriented dialogue tasks~\cite{1}.  Goal-oriented dialogue tasks are designed to help users complete specific tasks with clear task objectives~\cite{li2024review,deng2024towards}, such as bargaining, persuasion, negotiation, etc.~\cite{dilog2}. Although LLM significantly improves the performance of dialogue systems, their actual performance is heavily dependent on well-designed prompts. Due to the prevalence of a large amount of contextually relevant content in dialogue scenarios, such as negotiation, persuasion, and emotional support, and other complex interaction contexts, these tasks often require the system to have the ability to dynamically adjust dialogue strategies. Therefore, it is particularly important to create a policy planner in goal-orientated dialogue tasks to analyze the contextual needs in real time according to the dialogue process, and maintain the consistency of the goal-orientated dialogue and the effectiveness of the policy throughout the interaction. This planning mechanism is designed not only to improve dialogue quality, but also to enable the system to better adapt to complex and changing dialogue situations.

\begin{figure*}[t]
    \centering
    \includegraphics[width=\textwidth]{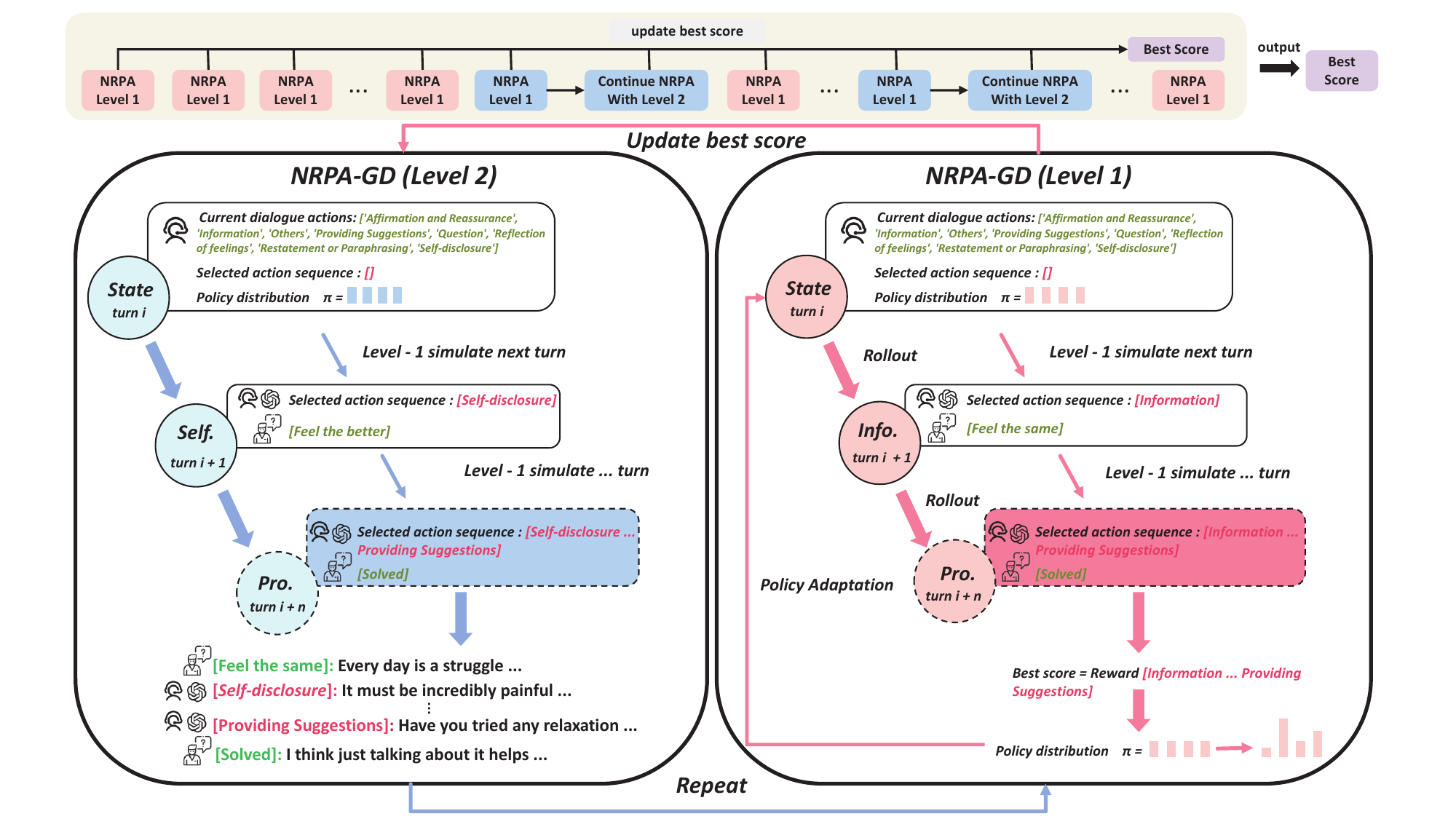} 
    \caption{The NRPA-GD system uses a nested architecture to implement recursive policy optimization through two levels. In level 2, simulations are performed by recursively calling level 1 to optimize the policy distribution based on the reward values generated by policy choices, and the updated policies are passed to the next round of simulation iterations.
    }
    \label{fig:nrpa_planner} 
\end{figure*}

Existing research has directly augmented LLM by designing heuristics or complex prompting processes, but such approaches are inefficient and have limited performance~\cite{ppdpp}. The subsequent emergence of policy planners combined with LLM, Yu et al.~\cite{gdpzero} use LLM to act as a prior policy, value function and complete tree search planning, resulting in improved performance, but still hardly satisfactory. To this end, He et al.~\cite{dpdp} propose a two-stage framework: first, a policy model is trained to quickly give policies based on context; then a policy planner performs fine-grained planning for unfamiliar scenarios. This scheme achieves the best results in goal-oriented dialogues, but overall efficiency is still low, because a large amount of dedicated data is required for training the policy model and retraining is needed every time when the dialogue scenario is changed.


With the development of the field of artificial intelligence in decision optimization, Monte Carlo Tree Search (MCTS) combined with neural networks demonstrated the power of MCTS in decision optimization\cite{alphago,mcts-5}. In recent years, researchers have begun to explore the combination of MCTS and LLM, where MCTS provides structured search capabilities while LLM contributes generative capabilities. Zhang et al.~\cite{mcts-1} proposed a self-training method based on process reward enhancement that automatically optimizes inference steps. In addition, Zheng et al.\cite{mcts-2} successfully solved complex optimization problems using MCTS heuristic rule generation. DeLorenzo et al.\cite{mcts-3} then combined the prospective search with the hardware design to generate high-performance hardware design codes. Gan et al.\cite{mcts-4} realized adaptive resource allocation for unsupervised tasks by dynamic collaboration of multiple intelligences. Inspired by MCTS as a dialog policy planner~\cite{gdpzero}, we treat the goal-oriented dialog task as a combinatorial optimization problem~\cite{poh}, which can be solved in a way similar to finding optimal paths under complex constraints. Many studies have shown that Nested Monte Carlo Search (NMCS)~\cite{nmcs} outperforms MCTS in a variety of optimization problems, and NMCS is particularly efficient in complex decision-making and optimization tasks~\cite{nmcs} because it finds the optimal solution by randomly sampling within a multilayer decision tree. The algorithm has been successfully applied to one-player games\cite{nmcs1}, two-player games~\cite{nmcs2}, and RNA back-folding problems~\cite{rna}. In addition, Nested Rollout Policy Adaptation (NRPA), an advanced variant of NMCS, has demonstrated improved performance in many domains, and NRPA further improves decision quality by optimizing the policy adaptation process~\cite{nrpa1}. Compared to NMCS, NRPA is particularly advantageous in scenarios that require fine-grained policy tuning and fast policy updates. NRPA has already achieved significant results on tasks such as Minimum Congestion Shortest Path Routing~\cite{cm}, Traveling Salesman Problem with Time Windows~\cite{tsp}, and Snake-in-the-Box~\cite{snake}, so the paradigm is also applicable to policy planning of dialogue scenarios.

Therefore, in this study, we propose a novel approach to plan the goal-oriented dialogue task based on NRPA. The NRPA is unique in that it employs a nested policy optimization mechanism, which achieves the global optimization of the dialogue policy through a multi-level policy evaluation and adjustment process. The method constructs complete dialogue trajectories in each dialogue simulation and iteratively updates the strategies based on these trajectories. This planning approach does not require pre-training of dialogue strategy models, but rather optimizes the strategy selection through online simulation and adaptive adjustment, thus effectively improving the goal attainment rate while maintaining dialogue fluency. Our comprehensive evaluation on multiple proactive dialogue tasks shows that NRPA-GD significantly outperforms existing policy planners on all metrics, and meets or surpasses that of pre-trained models for offline reinforcement learning. Our contributions can be summarized as follows.
\begin{itemize}
    \item We propose an online planner powered by LLMs for goal-oriented dialogue tasks based on NRPA, eliminating the need for any pre-training of specific dialogue strategy models.
    \item We introduce NRPA-GD as an efficient search methods, and our experimental results demonstrate that NRPA-GD significantly outperforms the MCTS-based approach while maintaining comparable or superior dialogue quality.
     \item Experiments demonstrate that NRPA-GD significantly outperforms existing policy planners in all metrics and achieves equivalent or even better performance than offline reinforcement learning training models.
\end{itemize}

\section{Related Work}
Dialog planning involves identifying appropriate strategies to guide the dialog before each dialog response. Existing prompt engineering approaches like Ask-an-Expert (AnE)~\cite{ane} prompt experts through predefined instructions to provide advice aligned with standard reasoning protocols in mental health, while the dialogue model learns from context and history which pieces of policy to adopt and which to discard. Proactive Chain-of-Thought~(ProCoT)~\cite{proactive} inserts an explicit goal-planning chain into the prompt so the model first infers its objective and then decides whether to probe further or refuse, revealing both the potential and the limitations of current large models in proactive conversation. 

When combining pre-trained models as planning strategies, the Plug-and-Play Dialogue Policy (PPDPP)~\cite{ppdpp} decomposes the dialogue policy into plug-and-play miniature models, trains them with supervised and reinforcement learning (RL) and significantly outperforms existing prompt or feedback-based approaches. Inspired by dual-process theory, Dual-Process Dialogue Planning (DPDP)~\cite{dpdp} splits the LLM dialogue policy into a fast intuitive path and a slow analytical path. An offline-RL lightweight policy handles familiar contexts; when the situation becomes complex or unknown, an MCTS depth search is triggered to refine the policy in real time. Tailored Strategic Planning (TRIP)~\cite{trip} integrates user-level traits into policy formulation, fusing user profiles and live feedback during planning through a population-based generalization framework that simulates multi-user distributions with evolutionary RL. User-Tailored Dialogue Policy Planning (UDP)~\cite{udp} uses a diffusion model to build user portraits on the fly, and predicts user feedback via a Brownian-bridge mechanism, and prioritizes hard samples with active learning, embedding individual traits into the policy loop to enhance system adaptability. Latent Dialogue Policy Planning (LDPP)~\cite{ldp} employs a variational autoencoder (VAE) to mine latent policies from real conversations and then trains a hierarchical policy planner offline in that latent space, avoiding simulation bias and validating the feasibility and efficiency of self-discovered policies from real data. However, relying directly on a large language model as the policy or patching it with MCTS remains brittle. PPDPP, DPDP, TRIP, UDP, and LDPP all depend on offline-RL strategies for planning. 

In addition, Yu et al.~\cite{gdpzero} propose Goal-oriented Dialogue Planning with Zero training (GDP-Zero), an MCTS-based policy planning method in which a single large language model simultaneously acts as a prior strategy, a value function, a user simulator, and a system model for unseen dialogue settings. Similarly, NRPA forgoes offline-RL entirely but refines policy from coarse to fine through an online multi-level recursive search~\cite{nrpa1} which achieve higher performance than MCTS in many domains. Therefore, we integrate LLMs into NRPA, to tackle the dialog tasks in this work.

\section{Method}
In this work, we propose a dialogue strategy planner based on the NRPA algorithm. This planner utilizes a zero-shot model training paradigm, performing multi-level exploration by prompting an LLM during decision-making. This process enables simulation of user-system responses, evaluation of current task progress, and optimization of the selection probability for the next dialogue act. As shown in Figure \ref{fig:nrpa_planner}, within the NRPA-GD framework, the simulation process is set to a maximum level of 2. This process operates through recursive calls, optimizing the policy distribution based on the reward values obtained from policy selection. The updated policy is then passed to the next round of simulation iterations. Using the ESConv dataset as an example, the simulation terminates when the user action is “Solved” or the maximum dialogue turns are reached, at which point the best score is updated.

\subsection{Problem Definition}
To introduce the NRPA method for dialogue policy planning, we first formulate the planning task as a Markov Decision Process (MDP). A dialogue consisting of $t$ turns between a user and a system can be represented as:
\begin{equation}
h_t = (u_0^{sys}, u_1^{usr}, u_1^{sys}, \dots, u_t^{usr}, u_t^{sys})
\end{equation}
where $u_i^{sys}$ is the system's response at turn $i$, and $u_i^{usr}$ is the user's utterance at turn $i$. Each system response $u_i^{sys}$ is associated with a specific dialogue act $a_i^{sys}$.
Similar to~\cite{yang2021, wang2020}, we define the task of planning the next $a^{sys}$ as an MDP problem $\langle S, A, T, R, \gamma \rangle$. The system's dialogue act $a_i^{sys}$ at turn $i$ represents an action $a_i \in A$. The corresponding dialogue history up to turn $i$,
\begin{equation}
s_i = (u_0^{sys}, u_1^{usr}, u_1^{sys}, \dots, u_{i-1}^{usr}, u_{i-1}^{sys})
\end{equation}
represents a state $s_i \in S$.
The transition function $T: S \times A \rightarrow S$ models the transition from state $s_i$ to $s_{i+1}$, which occurs after the system executes action $a_i$ to generate response $u_i^{sys}$ and then receives the user's subsequent utterance $u_{i+1}^{usr}$. The reward function $R(s)$ evaluates the quality of the final state $s$ of a simulated dialogue path. For instance, in an emotional support scenario, it measures whether the user's problem has been successfully resolved. We focus on evaluating the final state and implicitly incorporate discounting by applying a penalty to the dialogue length, thereby encouraging the model to achieve the goal in fewer turns. Specifically, if the dialogue is successfully solved, the reward value is 1, while a penalty of 0.001 times the dialogue turns is applied.

\subsection{NRPA Planner}
\begin{algorithm}[b!]
\caption{NRPA Playout($s$, $\pi$)}\label{alg:nrpa-playout}
\begin{algorithmic}[1]
\State $sequence \gets \emptyset$
\State $s \gets s_i$
\While{$s$ is not terminal}
    \State $z \gets \sum_{a' \in A} e^{\pi(a')}$
    \State Draw $a$ with probability $\frac{1}{z}e^{\pi(a)}$
    \State $u^{sys} \gets M_\theta(s \circ a)$
    \State $s \gets s \cup \{u^{sys}\}$
    \State append $a$ to $sequence$
\EndWhile
\State $score \gets R(s)$
\State \Return $(score, sequence)$
\end{algorithmic}
\end{algorithm}
The algorithm \ref{alg:nrpa-playout} is used to simulate the dialogue in state $s$ according to the current strategy $\pi$. The algorithm selects action $a$ using softmax probabilistic sampling, calls the large language model $M_\theta$ to generate system responses, and updates the state until the end of the dialogue to return the reward score and the sequence of actions.

\begin{algorithm}[t!]
\caption{NRPA Adapt($\pi$, $sequence$, $\alpha$, $s$)}\label{alg:nrpa-adapt}
\begin{algorithmic}[1]
\State $\pi' \gets \pi$
\State $s \gets s_i$

\For{$a$ in $sequence$}
    \State $z \gets \sum_{a' \in \mathcal{A}} e^{\pi(a')}$
    
    \For{$a' \in \mathcal{A}$}
        \State $\pi'(a') \gets \pi'(a') - \alpha \cdot \frac{1}{z}e^{\pi(a')}$
    \EndFor
    
    \State $\pi'(a) \gets \pi'(a) + \alpha$
    \State $s \gets play(s, a)$
\EndFor

\State \Return $\pi'$
\end{algorithmic}
\end{algorithm}

The algorithm \ref{alg:nrpa-adapt} implements adaptive updating of the strategy, receiving high-quality action sequences, and adjusting the weights of the strategy. For each action in the sequence, the mechanism of penalizing the global and rewarding the local is used to reduce the weights of all actions in proportion to the probability, and then significantly increase the weight of the current action, so that the strategy learns to optimize from the success experience. 

The algorithm \ref{alg:nrpa-main} implements multilevel strategy optimization, exploring dialogue paths, and selecting the best sequence update strategy through multiple playouts. The deeper levels of search provide more precise strategy guidance to the shallower levels, while the shallower levels are responsible for the broader exploration of the strategy space. This hierarchical optimization strategy enables NRPA-GD to efficiently discover high-quality dialogue strategies with limited computational resources, and achieves a novel combination of local fine-grained search and global strategy optimization.

\begin{algorithm}[b!]
\caption{NRPA($level$, $\pi$, $s$)}\label{alg:nrpa-main}
\begin{algorithmic}[1]
\Require Generative LLM $M_\theta$
\Require level $level$, Policy $\pi$
\Require Number of iterations $N$
\Require Action space $\alpha\in\mathcal{A}$

\If{$level = 0$}
    \State \Return \textsc{Playout}($s$, $\pi$)
\Else
    \State $bestScore \gets -\infty$
    \State $bestSequence \gets \emptyset$
    
    \For{$iteration = 1$ to $N$}
        \State $(score, sequence) \gets$ \textsc{NRPA}($level-1$, $\pi$, $s$)
        \If{$score > bestScore$}
            \State $bestScore \gets score$
            \State $bestSequence \gets sequence$
        \EndIf
        \State $\pi \gets$ \textsc{Adapt}($\pi$, $bestSequence$, $\alpha$, $s$)
    \EndFor
    \State \Return $(bestScore, bestSequence)$
\EndIf
\end{algorithmic}
\end{algorithm}

The effectiveness of MCTS depends heavily on the rollout policy used in the simulation phase, and past approaches have either used static uniform stochastic policies or relied on manually tailored heuristics for specific domains, which severely limits the efficiency of the search. The NRPA upgrades to a dynamic policy optimization framework based on the NMCS. Instead of explicitly unfolding the entire search tree node by node, the algorithm adjusts the weight of the rollout policy in real time through gradient ascent within each nested hierarchy, which exponentially amplifies the probability of generating a high return path. The probability of generating high-return paths is exponentially increased. Specifically, let the state of the first \(t\) step be \(s_t\), and let the set of legitimate actions be denoted as \(\mathcal{A}(s_t)\). We parameterize the strategy as a vector 
: $\pi \in \mathbf{R}^{|\mathcal{A}|}$, 
where the component \(\pi(a)\) corresponds directly to the weight of action \(a\). Given the optimal sequence of actions \((a_1,a_2,\dots,a_T)\) for a high score rollout, the following updates are performed for each step \(t\):

Calculate the softmax normalization factor.
\begin{equation} 
z = \sum_{a'\in\mathcal{A}} e^{\pi(a')}
\end{equation}

Calculate the probability of each action.
\begin{equation} 
P(a)=\frac{e^{\pi(a)}}{z}
\end{equation}

Update the weights for all actions \(a'\in\mathcal{A}\), and add an extra \(\alpha\) to the optimal action \(a\). 
\begin{equation} 
\begin{cases}
\pi(a')\gets\pi(a')-\alpha \cdot \frac{1}{z}e^{\pi(a')}, & \forall a' \in \mathcal{A} \\
\pi(a)\gets\pi(a)+\alpha
\end{cases}
\end{equation}

The net increase in weight of the optimal action \(a\) is \(\alpha - \alpha \cdot \frac{1}{z}e^{\pi(a)} = \alpha(1-P(a))\), and the net decrease in weight of the remaining actions is \(\alpha \cdot \frac{1}{z}e^{\pi(a')} = \alpha \cdot P(a')\), which transforms the original random simulation that was performed blindly into an adaptive sampling that continuously concentrates on the optimal direction.

\section{Experiments}
In this section, we introduce the experiment settings. 
\subsection{Datasets}
We evaluated the proposed framework on four active dialogue datasets: ESConv~\cite{esc} (Emotional Support Dialogue) containing 8 predefined actions; CIMA~\cite{cima} (Teaching and Learning Dialogues) containing 5 dialogue actions; and P4G~\cite{p4g} (Persuasion for Good). These three are all collaborative dialogue tasks (participants' goals are aligned), while CraigslistBargain~\cite{cb} (Price Negotiation) serves as a non-collaborative task (buyer seeks lowest price/seller seeks highest price) involving 11 buyer bargaining actions.


\begin{table*}[!t]
\centering
\label{table 1}
\begin{tabular}{@{}lcccc@{}}
\toprule
\multicolumn{1}{c}{} & \multicolumn{2}{c}{P4G} & \multicolumn{2}{c}{ESConv} \\
\cmidrule(r){2-3} \cmidrule(r){4-5}
Method & Run Time & Win Rate vs.\ ChatGPT & Run Time & Win Rate vs.\ ChatGPT \\
\midrule
GDP-Zero~\cite{gdpzero} & 636\,s & $58.66\%\pm2.73\%$ & 431\,s & $52.00\%\pm0.52\%$ \\
NRPA-GD (Level 1)  & \textbf{239\,s} & $68.61\%\pm1.84\%$ & \textbf{287\,s} & $63.40\%\pm0.52\%$ \\
NRPA-GD (Level 2)  & 1039\,s & $\textbf{77.49\%}\pm\textbf{2.25\%}$ & 1033\,s & $\textbf{74.50\%}\pm\textbf{2.26\%}$ \\
\bottomrule
\end{tabular}
\caption{Static evaluation with ChatGPT as backbone
and judge. Results are given as $\mu \pm \sigma$ repeated over three runs.}
\label{table 1}
\end{table*}

\begin{table*}[h]
\centering
\begin{tabular}{lcccccccccc}
\toprule
\multirow{2}{*}{Models} & \multicolumn{2}{c}{ESConv} & \multicolumn{2}{c}{CIMA} & \multicolumn{3}{c}{CraigslistBargain} \\
\cmidrule(lr){2-3} \cmidrule(lr){4-5} \cmidrule(lr){6-8}
 & AT$\downarrow$ & SR$\uparrow$ & AT$\downarrow$ & SR$\uparrow$ & AT$\downarrow$ & SR$\uparrow$ & SL$\uparrow$ \\
\midrule
DialoGPT~\cite{dialogpt}  &5.31& 0.7538 &5.43& 0.4956 & 6.73& 0.3245& 0.2012\\
\midrule
Standard~\cite{dpdp} & 5.10 & 0.7692 & 3.89 & 0.6903 & 6.47 & 0.3830 & 0.1588 \\
AnE~\cite{ane} & 4.76 & 0.8000 & 3.86 & 0.6549 & 5.91 & 0.4521 & 0.2608 \\
Proactive~\cite{proactive} & 5.08 & 0.7538 & 4.84 & 0.5310 & 5.80 & 0.5638 & 0.2489 \\
ProCoT~\cite{proactive} & 4.75 & 0.7923 & 4.58 & 0.5487 & 6.22 & 0.5319 & 0.2486 \\
ICL-AIF~\cite{iclaif} & 4.69 & 0.8079 & 4.19 & 0.6106 & 6.53 & 0.3617 & 0.1881 \\
PPDPP~\cite{ppdpp} & 4.56 & 0.8462 & 3.03 & 0.8407 & 5.62 & 0.6117 & 0.3376 \\
\midrule
DPDP (System 1)~\cite{dpdp} & 3.61 & 0.9000 & 2.24 & 0.9469 & 5.03 & 0.7447 & 0.4108 \\
-w/o PT & 4.22 & 0.8769 & 2.36 & 0.9292 & - & - & - \\
-w/o SPT & 3.97 & 0.8692 & 2.51 & 0.8938 & - & - & - \\
DPDP (System 2) & 2.13 & 0.9923 & 2.49 & 0.9735 & 2.78 & 0.9734 & 0.2728 \\
DPDP (System 1\&2) & \textbf{2.13} & 0.9923 & 2.28 & 0.9823 & - & - & - \\
\midrule
\midrule
NRPA-GD (Level 1) & 3.85 & \textbf{1.0000} & 1.08 & \textbf{1.0000} & 3.11 & \textbf{1.0000} & \textbf{0.6371} \\
NRPA-GD (Level 2) & 3.65 & \textbf{1.0000} & \textbf{1.03} & \textbf{1.0000} & \textbf{2.61} & \textbf{1.0000} & 0.5161 \\
\bottomrule
\end{tabular}
\caption{Automatic Evaluation Results with ChatGPT on ESConv, CIMA, and CraigslistBargain.}
\label{table 2}
\end{table*}

\begin{table*}[t]
\centering
\begin{tabular}{llccccccccc}
\toprule
\multirow{2}{*}{LLM} & \multirow{2}{*}{Level} & 
\multicolumn{2}{c}{ESConv} && \multicolumn{2}{c}{CIMA} && \multicolumn{3}{c}{CraigslistBargain} \\
\cmidrule(lr){3-4} \cmidrule(lr){6-7} \cmidrule(lr){9-11}
 & & AT$\downarrow$ & SR$\uparrow$ && AT$\downarrow$ & SR$\uparrow$ && AT$\downarrow$ & SR$\uparrow$ & SL$\uparrow$ \\
\midrule
\multirow{2}{*}{GPT-4o-mini} 
 &  1 & 5.28 & 0.9461 && 2.04 & \textbf{1.0000} && 3.89 & 0.9894 & 0.6326 \\
 &  2 & 4.17 & \textbf{1.0000} && 1.76 & \textbf{1.0000} && \textbf{2.72} & \textbf{1.0000} & \textbf{0.6422} \\
\midrule
\multirow{2}{*}{Llama3.1-8b} 
 &  1 & 3.98 & \textbf{1.0000} && 1.92 & \textbf{1.0000} && 3.54 & \textbf{1.0000} & 0.5166 \\
 &  2 & \textbf{3.65 }& \textbf{1.0000} && 1.69 & \textbf{1.0000} && 3.16 & \textbf{1.0000} & 0.4938 \\
\midrule
\multirow{2}{*}{Qwen3-0.6b} 
 &  1 & - & - && 1.22 & \textbf{1.0000} && 3.28 & 0.9894 & 0.6159 \\
 &  2 & - & - && \textbf{1.08} & \textbf{1.0000} && 3.17 & \textbf{1.0000} & 0.6307 \\
\bottomrule
\end{tabular}
\caption{Experimental results comparing LLM performance on ESConv, CIMA, and CraigslistBargain.}
\label{table 3}
\end{table*}

\subsection{Baselines}
Our aim is to demonstrate the effectiveness of Monte Carlo methods in a goal-oriented dialogue framework through NRPA. To this end, we systematically compare three classes of baseline methods. Dialogue models based on generalized fine-tuning techniques are represented by DialoGPT\cite{dialogpt}, a pre-trained dialogue generation model whose core functionality is to automatically generate natural, coherent and informative replies given a dialog context. Prompt-based engineering approaches like Standard Prompt\cite{dpdp} drive LLM generation of replies through base prompts; Proactive\cite{proactive} and ProCoT\cite{proactive} introduce explicit goal-planning chains in the prompts; Ask-an-Expert~\cite{ane} uses predefined prompts to model experts' standard reasoning strategies; ICL-AIF\cite{iclaif} generates textual feedback for zero-parameter updating contextual learning through model self-gaming; and GDPZero\cite{gdpzero} innovates by allowing large language models to assume multiple roles in tree search. Offline RL-based approaches include the fine-tuned small model scheme of PPDPP\cite{ppdpp}, and the dual processing mechanism of DPDP\cite{dpdp}, which combines offline RL training and real-time MCTS search optimization.

\subsection{Evaluation Metrics}
In the evaluation, we use three key metrics: Average Turns (AT) and Success Rate (SR). AT measures the efficiency of goal completion by calculating the average number of dialogue turns required to reach a goal. SR measures the effectiveness of goal completion by counting the percentage of successful goal completion within a predefined maximum number of turns. And SL (Sale-to-List Ratio) evaluates the buyer's transaction outcome. The higher the SL, the more the buyer benefits from the deal; if the deal fails, the SL is recorded as 0. The formula is defined as:\textit{ SL\% = (deal price - seller target price)/(buyer target price - seller target price) price)}. The same evaluation method of GDP-Zero was used in the P4G dataset, where we extracted the first 20 dialogues from P4G and generated a total of 154 rounds for evaluation~\cite{gdpzero}. Using chatgpt as a judge then prompts the ChatGPT judge to select the more persuasive response. In addition, we found bias in the direct assessment using LLM. Therefore, we also performed a human evaluation as a comparison. Based on the evaluation dimensions in \cite{dpdp}, we designed a multi-dimensional evaluation framework for different datasets. For the ESConv dataset, we compare the four dimensions of Suggestion (Sug.), Identification (Ide.), Comforting (Com.), and Overall (Ove.); for the CIMA dataset, we compare the four dimensions of Hint, Identification (Ide.), and Overall (Ove.); the P4G dataset focuses on Motivation (Motiv.), Persuasion (Pers.), Emotional (Emo.), and Overall (Ove.); and for the CraigslistBargain dataset it focuses on the Reasonableness (Rea.), Effectiveness (Eff.), Deal Success (Dea.) and Overall (Ove.). Among them, the evaluation dimensions of ESConv and CIMA directly adopt the criteria in \cite{dpdp}. In the evaluation process, each annotator needs to judge which one of the different levels of NRPA-GD performs better for each dimension and give a conclusion of win, lose, or tie. To ensure the objectivity and consistency of the evaluation, we provided detailed evaluation guidelines and examples for the annotators, and conducted labeling consistency tests before the formal evaluation. In the end, we summarized and averaged the assessment results of the three annotators to obtain a more reliable and objective assessment conclusion.

\section{Results and Analysis}
This section presents the experimental results and corresponding analysis.
\subsection{Static Evaluation}
Table \ref{table 1} compares the static GDP-Zero evaluation results with different nested levels of NRPA-GD on the P4G and ESConv datasets. The backbone model is gpt-4o-mini and the evaluator is gpt-3.5-turbo. GDP-Zero employs MCTS as the planner and adopts the best-reported setting from~\cite{gdpzero} with 50 simulations. NRPA-GD uniformly sets the number of single iterations at each level to 10 and introduces an early-stopping mechanism to curb the deepening of the layers with exponential time overhead caused by the deepening of the layers. The experimental results show that when the nesting level of NRPA-GD is 1, its planning response on P4G is preferred in up to 68. 61\%,  with much less time consumption compared to GDP-Zero. And in ESConv, the win rate for level 1 is up to 63. 40\%, with the lowest time cost. By further increasing the level to 2, the computational cost increases significantly,  but the best performance achieved on both datasets.

\subsection{Automatic Evaluation}
As shown in Table \ref{table 2}, the overall performance of NRPA-GD is accompanied by a tier improvement. In CIMA, the SR of both Level 1 and Level 2 reaches 100\%, while the AT decreases from the best 2.24 to 1.03, with a decrease of more than 54\%, which is better than the previous pre-trained policy models or the MCTS baseline. This suggests that a multilayer search can quickly target the optimal policy in tasks with small action space. In ESConv, although the AT (3.65) of Level-2 is still higher than the DPDP (2.13), the SR reaches 100\%, which is critical in emotional support scenarios where the problem is solved. In CraigslistBargain, Level-1 increased SL from 0.4108 to 0.6371 while maintaining SR at 100\% and AT at 3.11, and Level 2 decreased SL to 0.5161, but the AT was further decreased to 2.61, indicating that after increasing the number of simulations, NRPA-GD achieved a balance between AT, SR and SL. NRPA-GD achieves a success rate of 100\% in all three datasets by expanding the depth of search, and the focus on high-return dialogues can improve the surplus without sacrificing the deal completion rate.
\subsection{Performance on Different LLMs}
To further validate the effectiveness of the proposed framework under different model sizes, we use three different sizes of LLMs: gpt-4o-mini, llama-3.1-8b-chat, and qwen-3-0.6b. We evaluated the performance of NRPA-GD Level 1 and Level 2 respectively, and the results are shown in Table \ref{table 3}. With gpt- 4o-mini and llama-3.1-8b-chat, NRPA-GD shows a consistent performance improvement trend when the nesting level is increased, with both AT and SR improving and maintaining high returns on the CraigslistBargain dataset. This validates the stability and effectiveness of the NRPA-GD framework on medium- and large-scale models. For the qwen-3-0.6b model with the smallest parameter size, it is unable to complete the ESConv experimental task within the preset limit of dialogue rounds due to its parameter size limitation. However, the model still demonstrated the advantages of NRPA-GD on other datasets. In the CIMA data set, for both the AT and SR metrics of Level 2, the method achieves the desired high success rate performance with short rounds. And in the CraigslistBargain data set, the SL value of Level 2 also improves significantly to \textbf{0.6307}, which is comparable to the performance of gpt-4o-mini and gpt-3.5-turbo. It is evidenced that NRPA-GD is able to achieve the higher performance level than that of large parameter pre-trained models on specific tasks even with small parameter models.

\subsection{Human Evaluation}
Based on previous research \cite{esc}, LLM may lack rigorous criteria for persuasion components due to its possible preference for self-generated dialogues. In order to objectively assess the model performance, we randomly selected 50 sets of dialogues from each of the P4G and CIMA datasets for multidimensional human evaluation. During the evaluation process, we designed the corresponding evaluation dimensions for the specific task goals of different datasets. With an increase in search depth, the NRPA-GD in Figure \ref{fig:combined_results} shows a significant performance improvement on both datasets. In the P4G task, it can provide emotional support and promote donation behavior more effectively, and in the CIMA task, the model can provide students with more accurate learning hints. More importantly, at the methodology level, the NRPA-GD model demonstrates stronger guidance and flexibility, and can maintain consistent and high-quality performance in different dialogue scenarios.
\begin{figure}[!h]
    \centering
    \begin{subfigure}[b]{\linewidth}
        \includegraphics[width=\linewidth]{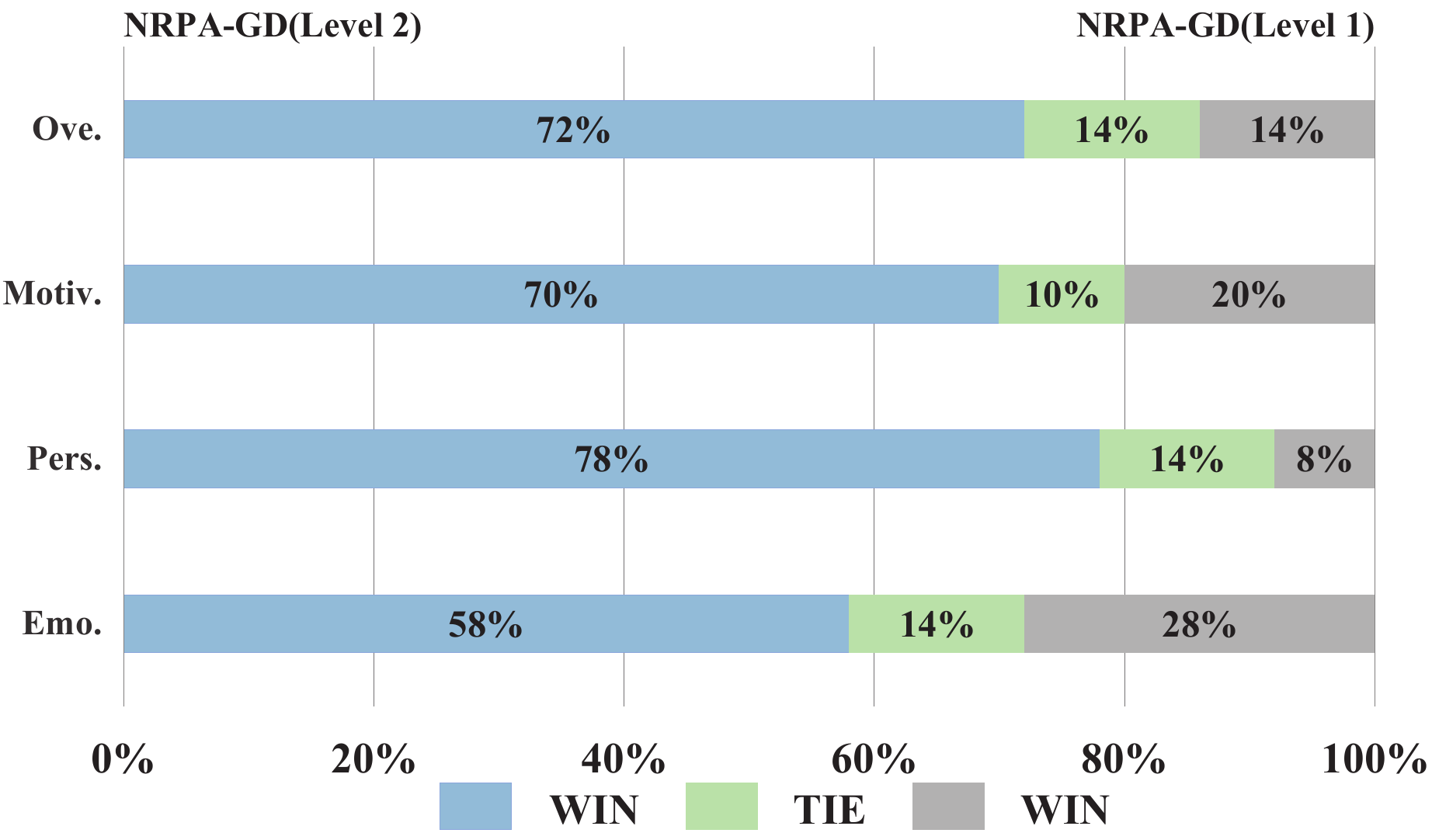}
        \caption{P4G.}
        \label{fig:P4G}
    \end{subfigure}
    \hfill
    \begin{subfigure}[b]{\linewidth}
        \includegraphics[width=\linewidth]{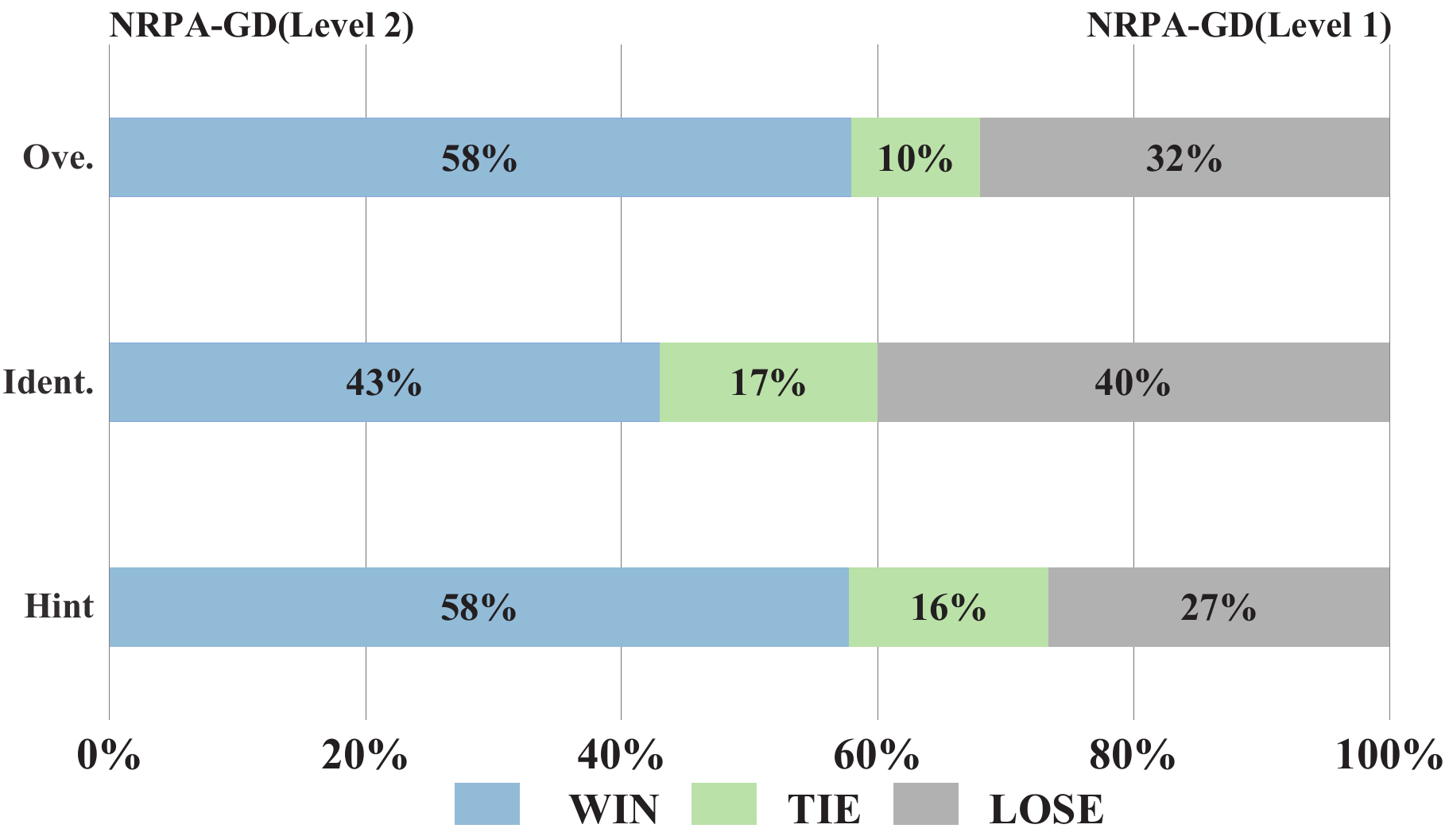}
        \caption{CIMA.}
        \label{fig:CIMA}
    \end{subfigure}
    \caption{Human evaluation results on P4G and CIMA.}
    \label{fig:combined_results}
\end{figure}



\begin{figure}[!ht]
    \centering
    \begin{subfigure}{0.48\linewidth}
        \includegraphics[width=\linewidth]{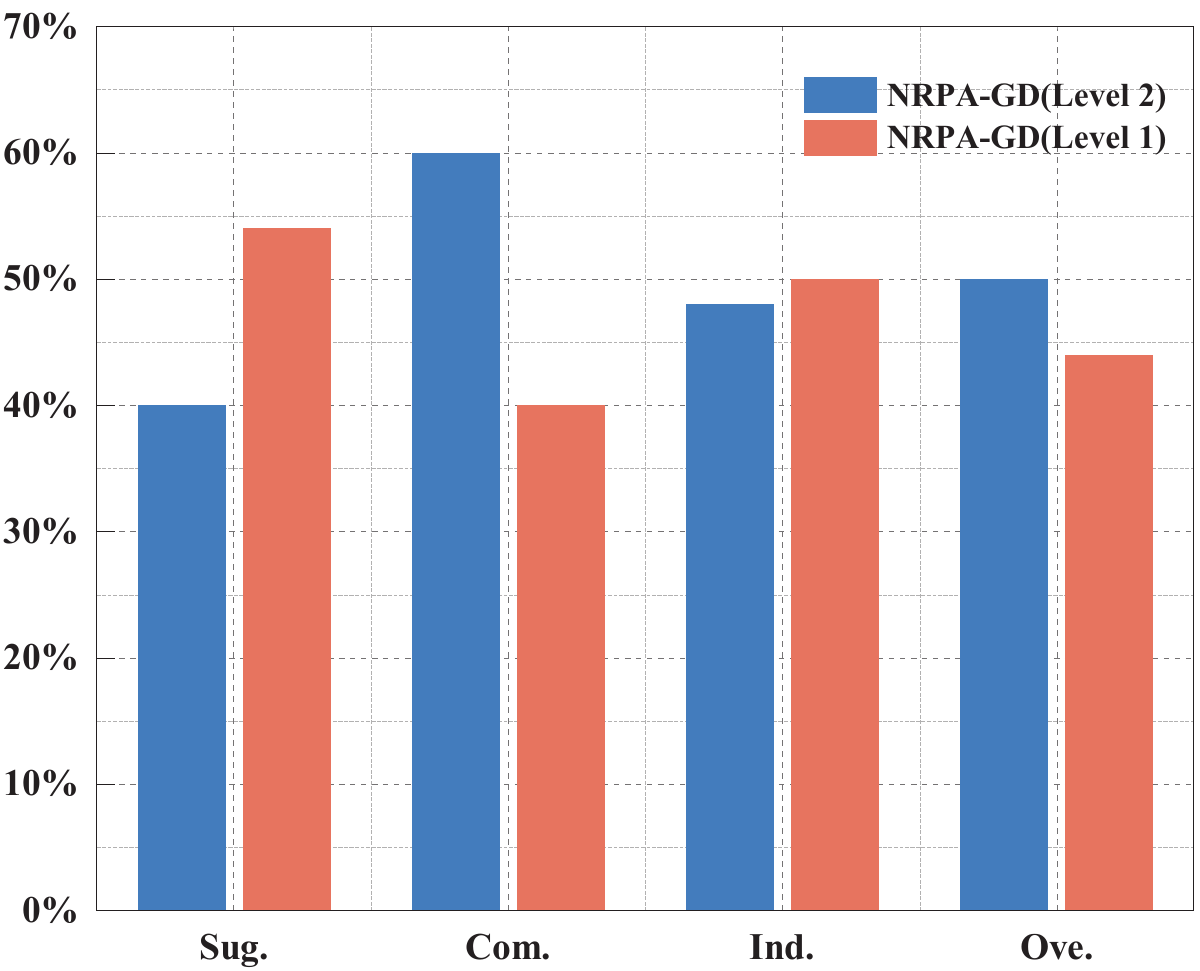}
        \caption{ESConv.}
        \label{fig:ESCONV}
    \end{subfigure}
    \hfill
    \begin{subfigure}{0.48\linewidth}
        \includegraphics[width=\linewidth]{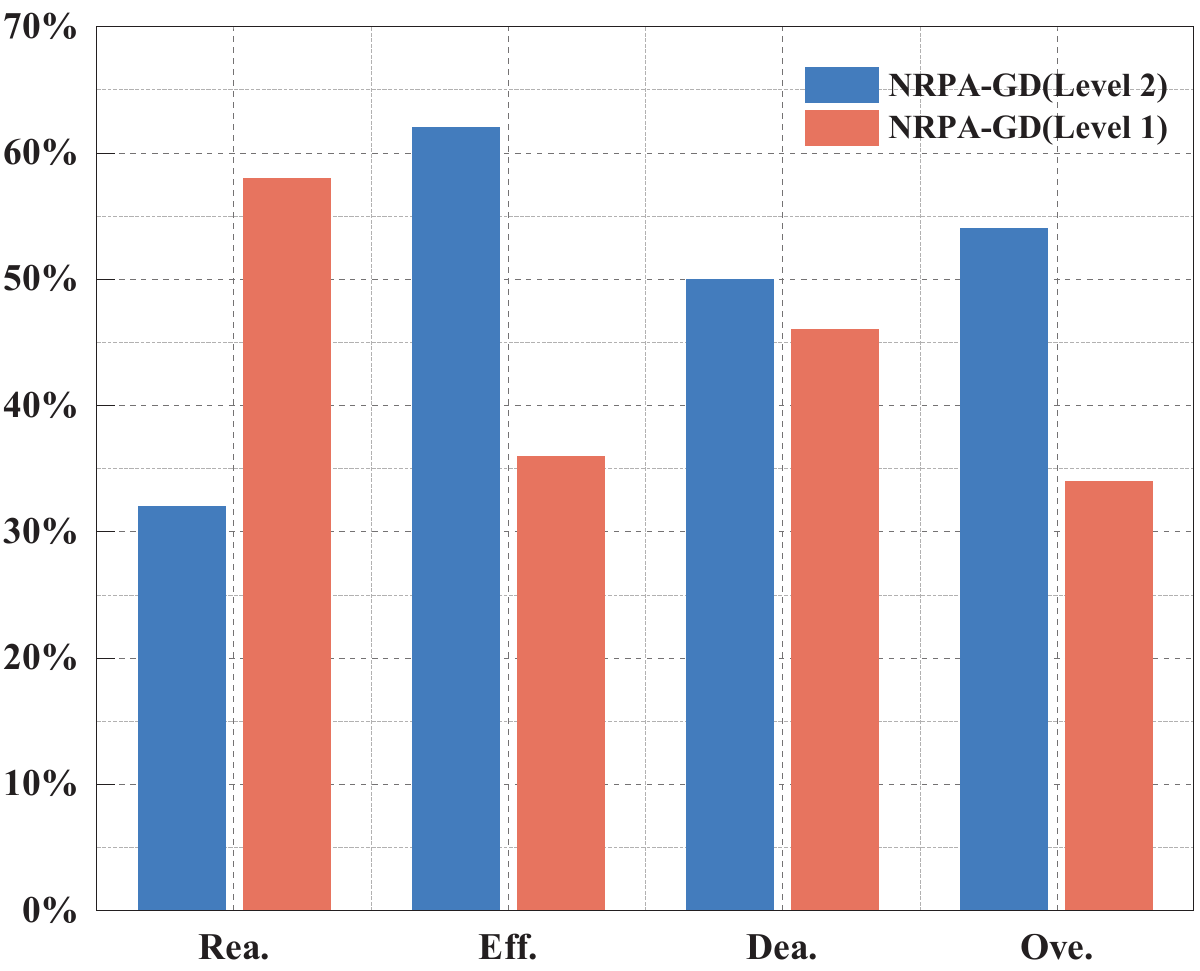}
        \caption{CraigslistBargain.}
        \label{fig:CB}
    \end{subfigure}
    \caption{Human evaluation results on ESConv and CraigslistBargain.}
    \label{fig:combined_results}
\end{figure}

Similarly, we manually evaluated 50 dialogues in each of the ESConv and CraigslistBargain datasets. The evaluation results in Figures \ref{fig:ESCONV} and \ref{fig:CB} reveal the performance of the model at different search depths. In the ESConv task, NRPA-GD (Level 1) tends to provide concrete and feasible suggestions, showing a strong problem solving orientation, while NRPA-GD (Level 2) focuses more on building emotional empathy with patients, and the two levels are comparable in terms of their ability to understand the emotional state of patients and identify core problems. In CraigslistBargain, the models exhibited different strategy preferences. nRPA-GD (Level 1) demonstrated more rational negotiation behavior, while NRPA-GD (Level 2) was superior in practical results, being more effective in reaching satisfactory negotiation outcomes and facilitating successful transactions, reflecting stronger goal orientation and execution. Based on superior performance, NRPA-GD (Level 2) outperforms Level 1 in the overall evaluation, validating the effectiveness of the deep search strategy in complex dialogue tasks. 

\section{Conclusion}
We propose NRPA-GD, an LLM-based online policy planning algorithm designed for goal-oriented dialogue systems. Moreover, it enables policy optimization without additional training. NRPA-GD's nested simulation mechanism enables dynamic exploration of multiple possible interaction paths during dialogue, while the adaptive update mechanism ensures that the policies can be optimized and adjusted based on real-time feedback. The experimental results show that the policies generated by NRPA-GD surpass the previous best system (DPDP), achieving stable and significant improvements in the evaluation metrics. NRPA-GD also surpasses ChatGPT even with an LLM with only 0.6B parameters, dramatically improving the intelligence of the dialogue system without increasing model parameters and training costs. For future work, it is necessary to explore more pruning methods to further improve the time efficiency of the proposed approach.

\section*{Acknowledgments}
The authors acknowledge the financial support from National Natural Science Foundation of China, No. 62236002, and  Hefei Key Science and Technology Special Projects under Grant 2024SZD006.


\bibliography{aaai2026}
\clearpage
\section{Appendix}
\subsection{Experimental Details}
In the experimental setup, we used the DPDP and GDP-Zero prompts and adopted the GDP-Zero code framework as the base implementation. To ensure the comparability and consistency of the experiments, we followed the standards of existing studies in dataset selection: the ESConv, CIMA, and CraigslistBargain datasets strictly follow the test set division of DPDP, while the P4G dataset is consistent with GDPZero. In terms of model configurations, both the dialogue system and the user model use gpt-3.5-turbo and gpt-4o-mini as the backbone models, and the key hyperparameter settings such as temperature parameters remain the same as the original configurations of DPDP and GDP-Zero to ensure the fairness and reproducibility of the experimental results. In addition, to verify the generalizability of our method on models of different scales, we also performed complementary experiments on smaller open-source models, including Llama-3.1-8B-Chat and Qwen3-0.6B, to evaluate the performance of NRPA-GD in resource-constrained environments.
\subsection{Nested Structure and Early Stopping Mechanism}
The NRPA algorithm employs a nested structure, with its iteration count and early stopping mechanism exhibiting different behaviors across different levels. At level 1, the algorithm performs five iterations, each of which calls level 0 for random playout. After finding the optimal dialogue sequence, the strategy is updated. At this point, the total number of simulations is five. When level 2 is reached, the top level still performs five iterations, but each iteration recursively calls level 1, which then performs its own five iterations. The early stopping strategy is applied independently at each level. When a certain level fails to find a better solution after continuous iterations, or when the number of dialogue rounds found reaches the threshold, that level will terminate the search early. The parameters are shown in Table \ref{tab:nrpa_parameters}.
\begin{table}[htbp]
\centering
\begin{tabular}{lcc}
\toprule
\textbf{Parameter} & \textbf{Value} & \textbf{Description} \\
\midrule
Level & 1 or 2 & NRPA level \\
Iterations & 10 & Number of iterations \\
Early\_stopping & 3 & Early stopping threshold\\
Min\_iterations & 3 & Minimum iterations \\
Max\_playout\_steps & 10 & Maximum turns \\
\bottomrule
\end{tabular}
\caption{Experimental parameters setting for NRPA-GD.}
\label{tab:nrpa_parameters}
\end{table}

\subsection{Human Evaluation Instruction}
For both the ESConv and CIMA datasets, following the DPDP setup, for ESConv we evaluate:

\begin{itemize}
    \item Suggestion: Which assistant provides more helpful suggestions for solving the problem?
    \item Comforting: Which assistant is more skilled at comforting you?
    \item Identification: Which assistant is more helpful in exploring and identifying the problem?
    \item Overall: Which assistant can better solve the patient's problem?
\end{itemize}

For CIMA, we measure two main aspects and an overall evaluation:

\begin{itemize}
    \item Hint: Which assistant provides more helpful hints for translating correctly?
    \item Identification: Which assistant is better able to identify students' translation errors?
    \item Overall: Which assistant can better teach the student?
\end{itemize}
For P4G, we assess the following key dimensions of persuasive donation appeals:

\begin{itemize}
    \item Donation Motivation: Which better motivates donation consideration
    \item Persuasive Argument: Which has more convincing arguments
    \item Emotional Resonance: Which connects better emotionally
    \item Overall: Which is better for convincing donation
\end{itemize}

For CraigslistBargain, we evaluate multiple aspects of negotiation performance in bargaining scenarios:

\begin{itemize}
    \item Reasonableness: Which assistant shows more reasonable negotiation behavior?
    \item Effectiveness: Which assistant is more effective at achieving a good outcome?
    \item Deal Success: Which assistant is more effective at reaching a successful deal?
    \item Overall: Which assistant is better overall for a bargaining scenario?
\end{itemize}
\subsection{Prompting Details}
We present the prompting details in
our implementation. All the prompts we used are
consistent with those in GDP-Zero and DPDP. In the P4G dialogue, it is through gradual loading that true conversation is achieved.
\subsection{Assistant Simulation} 
First, we will explain the design of role-playing prompts in the dialogue system, which guides subsequent dialogue action through dialogue strategy prompts [action].
\subsubsection{ESConv}
In emotional support dialogues, the assistant acts as a therapist, helping patients deal with emotional issues and personal difficulties. The dialogue begins with the user describing [situation], providing a specific context for the entire dialogue. Detailed prompt is presented in Table \ref{tab:esconv-prompts}.
\subsubsection{CIMA}
In the tutoring dialogue system, the assistant takes on the role of a teacher, specializing in guiding students through English-to-Chinese translation exercises. Each dialogue begins with a translation task marked as [exercise], and includes specific difficulties described by the student through [situation] tags, providing a unique background setting for the dialogue. Detailed prompt is presented in Table \ref{tab:cima-prompts}.
\subsubsection{CraigslistBargain}
In the negotiation dialogue, the assistant plays the role of the buyer and negotiates the price of the item with the seller. Each negotiation scenario includes [item name] and [item description] to provide a complete negotiation context. The buyer sets a clear target price [buyer target price], while the starting price for negotiation is set at the original price of the item [seller target price]. Detailed prompt is presented in Table \ref{tab:craigslist-prompts}.
\subsection{User Simulation}
We will explain the design of role-playing prompts used to guide large language models (LLMs) in simulating user behavior. This design ensures that simulated users generate responses based solely on dialogue history by excluding dialogue strategy prompts.
\subsubsection{ESConv}
In emotional support dialogue, the assistant plays the role of a patient who needs the therapist's help. The prompt template contains two key parameters: [emotion type] labels the patient's emotional state, and [problem type] defines the specific problem category that needs to be solved.
\subsubsection{CIMA}
In the tutoring dialogue, the assistant plays the role of a student learning English-Italian translation skills. Given that LLMs have excellent translation capabilities, we specifically ask the model to temporarily forget the standard translation of the current discussion exercise.
\subsubsection{CraigslistBargain}
In the negotiation dialogues, the assistant takes on the role of the seller, bargaining with the buyer for a higher item price.
\subsection{Static Evaluation}
In our static evaluation, we prompted ChatGPT to choose the better of two generated responses. Given two responses, we first provided the relevant task context and dialogue history, then asked ChatGPT which of the two responses would be more effective. For each evaluation, we conducted 5 sample tests and used a majority voting mechanism. Detailed prompt is presented in Table \ref{tab:persuasion-prompts} and \ref{tab:esconv-prompts1}.

\clearpage
\begin{table*}[!t]
\centering
\begin{tabular}{lp{0.8\textwidth}}
\toprule
System & Now enter the role-playing mode. In the following conversation, you will play as a therapist in a counseling conversation with a patient. \\
\midrule
User & You are the therapist who is trying to help the patient reduce their emotional distress and help them understand and work through the challenges. Please reply with only one short and succinct sentence. [action] Are you ready to play the game? \\
Assistant & Yes, I'm ready to play the game! \\
\midrule
User & [situation] \\
\bottomrule
\end{tabular}
\caption{Prompts for assistant simulation on ESConv.}
\label{tab:esconv-prompts}
\end{table*}

\begin{table*}[!h]
\centering
\begin{tabular}{lp{0.8\textwidth}}
\toprule
System & Now enter the role-playing mode. In the following conversation, you will play as a teacher in a tutoring conversation with a student. \\
\midrule
User & You are the teacher who is trying to teach the student to translate [exercise] into Italian. Please reply with only one short and succinct sentence. Please do not tell the student the answer or ask the student about other exercises. [action] Now ask me an exercise. \\
Assistant & Please translate [exercise] into Italian. \\
\midrule
User & [situation] \\
\bottomrule
\end{tabular}
\caption{Prompts for assistant simulation on CIMA.}
\label{tab:cima-prompts}
\end{table*}

\begin{table*}[!h]
\centering
\begin{tabular}{lp{0.8\textwidth}}
\toprule
System & Now enter the role-playing mode. In the following conversation, you will play as a buyer in a price bargaining game. \\
\midrule
User & You are the buyer who is trying to buy the [item name] with the price of [buyer target price]. Product description: [item description] Please reply with only one short and succinct sentence. [action] Now start the game. \\
Assistant & Hi, how much is the [item name]? \\
\midrule
User & Hi, this is a good [item name] and its price is [seller target price]. \\
\bottomrule
\end{tabular}
\caption{Prompts for assistant simulation on CraigslistBargain.}
\label{tab:craigslist-prompts}
\end{table*}

\begin{table*}[!h]
\centering
\begin{tabular}{lp{0.8\textwidth}}
\toprule
& The following is background information about Save the Children. \\
& Save the Children is head-quartered in London, and they work to help fight poverty around the world. \\
& Children need help in developing countries and war zones. Small donations like \$1 or \$2 go a long way to help. The following is a conversation between a Persuader and a Persuadee about a charity called Save the Children. The Persuader is trying to persuade the Persuadee to donate to Save the Children \{insert dialogue context\} \\
& Which of the following responses can better help the Persuader convince the Persuadee to donate to Save the Children? Why? \\
& A. Persuader: \{insert response a\} \\
& B. Persuader: \{insert response b\} \\
& C. Can't tell. \\
& Your can choose from either A, B, or C. \\
& Your choice: \\
\bottomrule
\end{tabular}
\caption{Static evaluation Prompts on P4G.}
\label{tab:persuasion-prompts}
\end{table*}

\begin{table*}[!t]
\centering
\begin{tabular}{lp{0.8\textwidth}}
\toprule
& The following is a conversation between a Therapist and a Patient in an emotional support session. \\
& The patient is seeking help for emotional distress, and the therapist aims to provide effective psychological support to alleviate the patient's concerns or improve their emotional well-being. \\
& Conversation Context: \\
& \{context\} \\
& Which of the following therapist responses would more effectively help the patient address their emotional struggles or feel supported? \\
& A. Therapist: \{resp\_a\} \\
& B. Therapist: \{resp\_b\} \\
& C. Hard to tell \\
& Choose the best option (A, B, or C). \\
& Your choice: \\
\bottomrule
\end{tabular}
\caption{Static evaluation Prompts on ESConv.}
\label{tab:esconv-prompts1}
\end{table*}
\end{document}